# An Embedding-Based Grocery Search Model at Instacart

Yuqing Xie[*][†]
yuqing.xie@uwaterloo.ca
University of Waterloo
Canada

Taesik Na[*]
taesik.na@instacart.com
Instacart
USA

Xiao Xiao
xiao.xiao@instacart.com
Instacart
USA

Saurav Manchanda
saurav.manchanda@instacart.com
Instacart
USA

Young Rao
young.rao@instacart.com
Instacart
USA

Zhihong Xu
zhihong.xu@instacart.com
Instacart
USA

Guanghua Shu
guanghua.shu@instacart.com
Instacart
USA

Esther Vasiete
esther.vasiete@instacart.com
Instacart
USA

Tejaswi Tenneti
tejaswi.tenneti@instacart.com
Instacart
USA

Haixun Wang
haixun.wang@instacart.com
Instacart
USA

## ABSTRACT

The key to e-commerce search is how to best utilize the large yet noisy log data. In this paper, we present our embedding-based model for grocery search at Instacart. The system learns query and product representations with a two-tower transformer-based encoder architecture. To tackle the cold-start problem, we focus on content-based features. To train the model efficiently on noisy data, we propose a self-adversarial learning method and a cascade training method. On an offline human evaluation dataset, we achieve 10% relative improvement in RECALL@20, and for online A/B testing, we achieve 4.1% cart-adds per search (CAPS) and 1.5% gross merchandise value (GMV) improvement. We describe how we train and deploy the embedding based search model and give a detailed analysis of the effectiveness of our method.

## 1 INTRODUCTION

Instacart provides grocery shopping services for customers who order groceries from retailers, while shopping is done by individual shoppers. Millions of products are listed for everyday grocery purchases. A high-quality search system is critical to system efficiency and user experience. We build a search embedding model to learn the semantic relationship between queries and products. The model produces vector representations for queries and products, with the goal of minimizing/maximizing the vector distance of related/irrelevant <query, product> pairs.

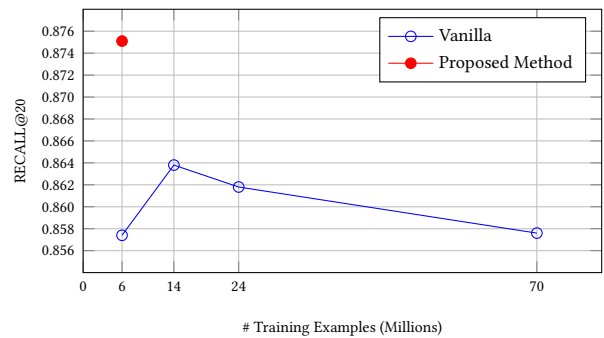

Figure 1: Vanilla training with a larger dataset doesn't guarantee better performance due to the noise in the data. Our methods utilize the knowledge learned from the vanilla training by performing cascade training with self-adversarial loss.

With the development of embedding-based retrieval models, much work has been done on embedding models for e-commerce. Amazon[12], Taobao[9] and JD [27] shared their embedding-based product search systems. We follow this line of research but utilize pre-trained transformer models [3] unlike previous approaches.

The first problem we tackle is cold-start. New products or unseen queries historically don't perform well due to the lack of user engagements. Furthermore, according to our user research, new users tend to send unseen queries because they are not familiar with the e-commerce search engine. It's critical to show high-quality results for these queries so that we can retain these new users. Additionally, retailers expect good ranking results for newly introduced products. We can manually boost some of the new products to meet business needs, but manual boosting doesn't scale at all.

Previous works either include historical features, such as clicks, cart adds (conversions), purchases in the model input, or highly rely on these data in offline evaluation. If a system takes historical features as inputs, it will favor the products that have higher engagement. Continuous training on a regular cadence using these historical features will make the cold-start problem even worse, entering a self-fulfilling circle. Similar behavior will occur if the offline evaluation highly relies on the user engagement log.

---
[*]Both authors contributed equally to this work.
[†]This work was performed when the author was at Instacart.

To address this challenge, we propose to use only content-based features such as query text itself and product information including product name, brand name, categories and attributes from our catalog. This way, the model will focus on semantic relationships rather than historical engagement relationships. We also propose to use permutations of product information as synthetic queries for a product, and include those in the training dataset. This helps the model generalize on cold-start products when given attribute related queries. For offline evaluation, instead of relying on user engagement log data, we propose to evaluate the system on a scaled human annotated dataset, which contains a set of hot and cold products.

The second problem we tackle is noisy data. We follow previous approaches [9, 10], and use the user engagement data to collect positive query-product examples. We treat converted [1] products for queries as positive training samples. However, for grocery search, user engagement data could be particularly noisy. People come to Instacart to shop for multiple items. When irrelevant items show up in the search results, people may add these items to their cart just because they happen to need them regardless of the original search intent. When we take these conversions as training examples, it results in false positive labels. Another source of noise is that users may have very different tolerances for search results: Some users may add items that don't fully satisfy their search intent, while others may be very strict due to allergies or religious reasons. Therefore, conversions can imply a wide range of relevance, ranging from somewhat relevant to strongly relevant.

To show the effect of noise in the training set, we trained embedding based search models with different sizes of training data. As the training set gets larger, we potentially add more relevant but cold examples at the expense of including more noisy examples. As shown in Figure 1, the performance of the model decreases after a point. This shows training with more data doesn't necessarily give better model performance due to the inclusion of noise.

To address the noisy data problem, we propose cascade training: We first warm-up the model with a noisy but large dataset. This way, we transfer the pretrained model's knowledge to the grocery domain by letting the model see as much data as possible. And then we train the model on a smaller yet high quality data. In this step, we best utilize learned knowledge to achieve better performance by training on the high quality data. We also show that cross architecture training, first training the two-tower shared encoder and then training with un-tied parameters, is useful in cascade training. Shared encoder allows the model to learn common grocery knowledge, and two towers with un-tied parameters allows the model to best fine-tune for each expertise (query and product knowledge).

---

[1] In our case, a user converts a product in a search query means the user adds the product presented in the search results for the query to cart.

We also propose a self-adversarial negative sampling/re-weighting method to select informative negative samples. Arguably, the straightforward approach to generate negative samples is to draw them from a uniform distribution. Such uniform negative sampling is not efficient enough because many negative samples can be easily distinguished from the positive ones after a few training epochs. The proposed self-adversarial learning method adjusts the loss term weights for individual negative samples according to their scores in the current training step. This way, the model will be punished more, when it predicts high scores for the negative query product pairs, resulting in more efficient training. The main contributions of this paper are as follows:

- We propose cross architecture cascade training, and a self-adversarial negative sampling/re-weighting method, to efficiently train search embedding models on noisy data.
- We propose to only use content-based features, and add catalog synthesis data to help with the cold-start problem.
- The proposed model outperforms the previous baseline by 10% in RECALL@20 on an offline human evaluation dataset. In online A/B testing, the new system improves CAPS and GMV by 4.1% and 1.5% respectively.
- We share our knowledge in model architecture, training details, system deployment and downstream applications. We also conduct extensive ablation study and analysis to share our insights.

## 2 RELATED WORK
### 2.1 Embedding Based Search Systems
Many embedding based retrieval approaches, such as [11, 24] have been introduced to solve e-commerce search problems. Facebook systematically shared their knowledge in modeling, serving and full-stack optimization in [5, 10]. Taobao[9] and JD [27] shared their exploration in personalized product search systems. We also apply embedding based methods for product search, while we focus more on building a semantic relevance model.

### 2.2 Effective Negative Sampling Strategies.
Training large models often relies on good optimization strategies, of which negative sample strategies are one of the most important for representation learning. Yih et al. [26] introduces the in-batch negative sampling, taking other samples within a mini-batch as negative samples. This method doesn't require explicitly labeled negative data and can reduce computation by re-using other in-batch sample representations. However, uniformly selecting negative samples is not the optimal choice. Wieting et al. [23] proposes self-adversarial training, which takes the in-batch negative with the highest score as a hard negative, thus, achieving improved results for paraphrastic sentence embeddings. Later, self-adversarial negative sampling is proven useful for relation-aware graph attention models[13] and knowledge graph (KG) embeddings [18]. In this paper, we introduce a self-adversarial negative sampling strategy to help improve the training efficiency with noisy data.

### 2.3 Pre-trained Models and Transfer Learning
Over the past few years, pre-trained language models have large impact on natural language processing. Models such as BERT[3],

XLNet[25], T5[14], BART[8], GPT-3[1] effectively transfer the knowledge in large general domain corpus to target domains or downstream tasks. However, these large models can hardly be used in industry due to the high computation cost. DistilBERT[17], TinyBERT[6] and MiniLM[21] propose to distill from large pre-trained models into smaller models. To take advantage of both the generalization ability and the compressed size, we start from these distilled pre-trained models.

### 2.4 Curriculum Learning

We propose a cascade training method, which uses different training sets in two steps. This is similar to curriculum learning, a strategy that trains a model from easier to harder data, imitating the learning order in human curricula[22]. However, we focus more on best utilizing noisy data through scheduling according to data quality instead of difficulty.

## 3 TRAINING DATASET
### 3.1 Cascade Training

We collect conversions as positive training query-product samples. We gather one years worth of user engagement log, and only keep the query-product pairs that are converted by at least two users. As shown in Figure 1, including more pairs does not necessarily improve the model performance due to data noise. To filter out samples that are likely noise, we rank the products by the conversion rate in descending order and remove ones below a threshold. On one hand, we need to provide the model with more accurate relevance signals, so that it can learn a better relevance representation. On the other hand, we also need to train the model on as many unique queries or products as possible to transfer the pre-trained model's ability to our domain.

To meet both requirements, we propose cascade training: Starting from a pre-trained model, we first train the model on a warm-up dataset to transfer the model to the grocery domain. Then we train it on a cascade training dataset to achieve better performance. The entire training procedure is summarized in Figure 2.

**The warm-up training step:** We first collect a relatively large *warm-up dataset* with lower conversion rate threshold. For a query $q_i$, we select top $k_{q_i}\%$[2] products as positive pairs. To balance the training examples, we apply smaller/larger $k_{q_i}\%$ s for queries with higher/lower frequencies in the log. We finally gathered 14 million unique query-product pairs for the warm-up dataset.

**The cascade training step:** We then collect a smaller but less noisy *cascade training dataset*. We repeatedly include positive pairs with higher conversion rate threshold. We start with threshold $k_{q_i}^0\%$, which is much smaller than $k_{q_i}\%$ in the *warm-up dataset*, so that there are less noisy conversion signals. At step $n$, we add all the products ranked above top $k_{q_i}^n\%$ into the dataset. Then we decrease $k_{q_i}^n = k_{q_i}^{n-1} * \theta$ and repeatedly add sample pairs until there are no positive pairs. We finally gathered 6 million training samples for cascade training. Details of constructing the *cascade training dataset* can be found in Algorithm 1.

---

**Algorithm 1** Cascade Training Dataset Construction

$Q \leftarrow$ the query set
$C_{<q_i,p_j>} \leftarrow$ the number of conversions of $<q_i, p_j>$,
$P_{q_i} \leftarrow$ the converted products of $q_i$, sorted by $C_{<q_i,p_j>}$,
$n \leftarrow 0, D_{cascade} \leftarrow \emptyset, k_{q_i} \leftarrow k_{q_i}^0, \theta \leftarrow 0.5$,
**repeat**
$\quad D' \leftarrow \emptyset$
$\quad$**for** $q_i \in Q$ **do**
$\quad\quad D'$ append $\{<q_i, p_j> | p_j \in P_{q_i}, j \leq len(P_{q_i}) * k_{q_i}^n\%\}$,
$\quad\quad k_{q_i}^{n+1} \leftarrow k_{q_i}^n * \theta$
$\quad$**end for**
$\quad n \leftarrow n + 1$
$\quad D_{cascade} \leftarrow D_{cascade} + D'$,
**until** $D' = \emptyset$
**return** $D_{cascade}$

---

### 3.2 Augmentation with Catalog Data

Our users often combine different product keywords to create the queries. Suppose a user wants to by a product that has product name "Organic 2% Reduced Fat Milk", product brand name "GreenWise", category "Milk", and attributes "organic, kosher, gluten free", the user might send one of the following queries: *"GreenWise Milk", "Organic", "Organic 2% Reduced Fat Milk", "GreenWise"*. Our catalog contains these product meta data.[3] So we can apply this heuristic to augment the training dataset to help address the cold start problem.

We synthesize positive queries using the combination of one or more product feature sources, including product name, product attributes, categories and brand names. We randomly sample synthetic data, and merge these with the dataset created from the user engagement data. Although the synthetic data can be clean, we keep the ratio of the synthetic data relatively low because it changes the actual search distribution.

### 3.3 Features

We use the query text as the input of the query encoder. For the product encoder, we input the concatenation of product name, brand name, size information, categories and attributes. In this way, the model can only capture semantic relationships and generalize to cold start products or queries. We use special tokens to explicitly indicate the start of different features. This provides a grammar rule so that a model can easily identify different features.

An example of the query and product features would be *"[QRY] milk"* and *"[PN] Organic 2% Reduced Fat Milk [PBN] GreenWise [PCS] Milk [PAS] organic, kosher, gluten free"* respectively. In the above examples, we use [QRY], [PN], [PBN], [PCS] and [PAS] tokens to indicate the next token is a query, product name, product brand name, product categories and product attributes respectively.

## 4 NEGATIVE SAMPLING STRATEGIES

In grocery search, many products can satisfy the same query intent but the users just favor some of the them. So, unlike in [5], we do not collect products that are presented in the search results but not

---

[2]Using $k_{q_i}\%$ = 100 % means we include all the conversions for query $q_i$.

[3]We use hierarchical taxonomy, and map products into one of taxonomy nodes.

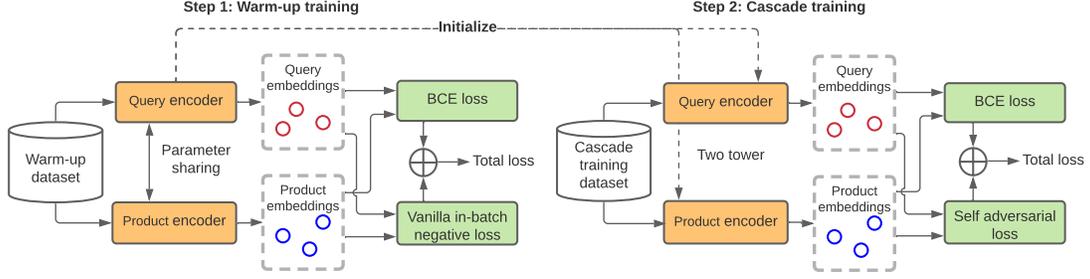

Figure 2: Cascade training overview. In step 1, we train a shared encoder model on the warm-up dataset. In step 2, we train the two encoders with un-tied parameters on the cascade training set with self-adversarial online negative sample re-weighting.

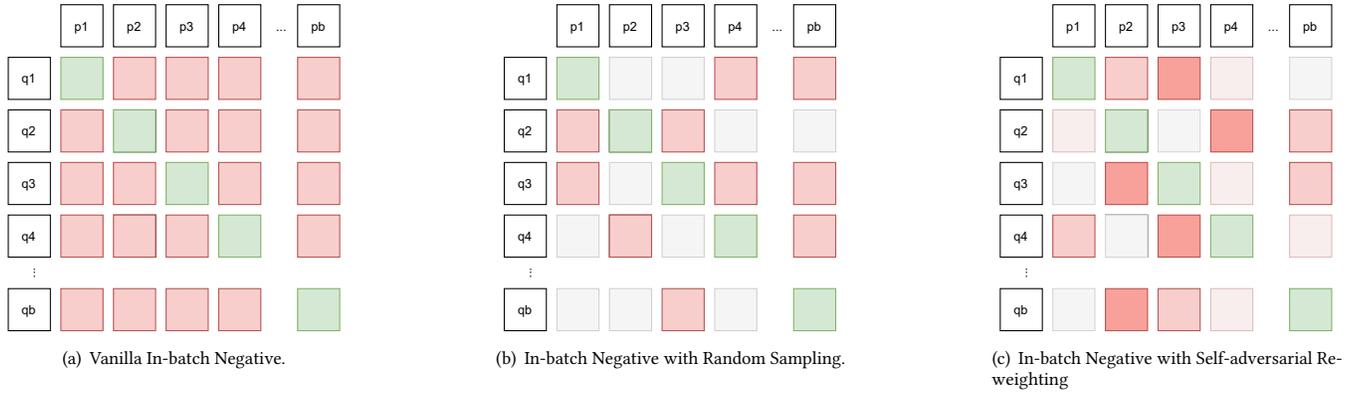

(a) Vanilla In-batch Negative.

(b) In-batch Negative with Random Sampling.

(c) In-batch Negative with Self-adversarial Re-weighting

Figure 3: Figure of different in-batch negative sampling. Positive samples are colored in green and negative ones are colored in red. 3(a) shows we include all the other in-batch samples as negative samples; 3(b) shows we randomly select negative samples in the batch, dropped ones are colored in grey. 3(c) shows the score re-weighted for adversarial negative samples, different levels of red represents the different weight applied to the samples.

converted as hard negatives. We only collect converted pairs and use in-batch negative samplings during training. [4]

### 4.1 Vanilla In-Batch Negatives

The vanilla in-batch negative strategy is to take all other in-batch products as negative samples, following Yih et al. [26]. Figure 3(a) demonstrates the samples we included in a mini-batch. For a training batch, we have $(q_1, q_2, ..., q_b)$ and $(p_1, p_2, ..., p_b)$, where $b$ represents the batch size, and $q_i$ and $p_i$ are query and product embeddings for a positive training example $i$ respectively. We take the dot products of the embeddings followed by a sigmoid activation $\sigma(\cdot)$ as the relevance score, and then we apply binary cross entropy loss $BCELoss(\cdot)$[5] to train the model. The objective is defined by: $L_{rel} = L_{pos} + \lambda_{neg} * L_{neg}$, where $L_{pos,neg}$ are binary cross entropy losses, and $\lambda_{neg}$ is a hyper-parameter to adjust negative loss weight.

---
[4]We experimented on adding hard negative products (presented in search results but not converted). But this gives worse performance on our human evaluation dataset due to the bias we manually introduce.
[5]We also experiment with other loss functions, such as Cosine similarity, Marginal loss, however, the BCE gives best performance.

### 4.2 Sampled In-Batch Negatives

For a certain query, instead of adding all negative loss terms into the final loss, we also try uniform sampling of negative products $S_i$ (Figure 3(b)). Each negative example $i$ has a sampling probability of $P(p_j \in S_i, i \neq j) = \frac{1}{b-1}$. The loss function becomes:

$$L_{rel} = L_{pos} + \lambda_{neg} * L_{sampled\_neg}, \quad (1)$$

$$L_{sampled\_neg} = \sum_i \sum_{p_j \in S_i} BCELoss(\sigma(q_i \cdot p_j), 0). \quad (2)$$

### 4.3 Self-Adversarial Negatives

Uniform negative sampling is not an optimal strategy: many negative samples can be easily distinguished after a few epochs of training. We hope the model can focus more on the difficulty samples. The difficulty of a certain negative sample to the current model can be measured by the model's prediction score. We adopt self-adversarial negative product sampling or re-weighting, shown in Figure 3(c). We take the prediction score of the current model for the negative sample as a weight term. We then apply either loss re-weighting or re-sampling based on the weight.

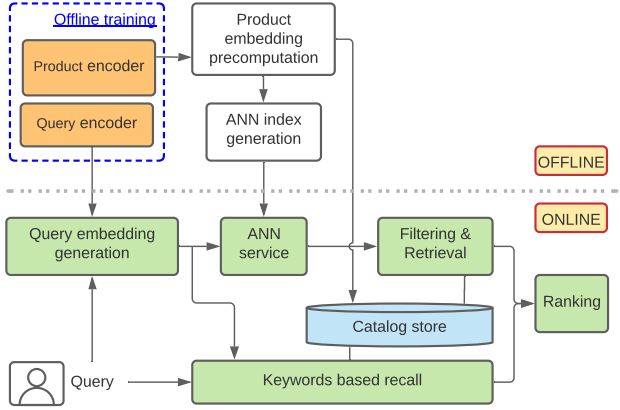

Figure 4: Overall system architecture that utilizes the embedding model for retrieval and ranking.

**Self-Adversarial Negative Sample Re-weighting:** We first compute the dot product of the query and product embedding vectors and pass the score through an activation function $act()$ to get the relevance score. We multiply each sample's loss with its relevance score. Take the BCELoss version for example, the new loss function becomes:

$$L_{rel} = L_{pos} + \lambda_{neg} * L_{self\_adv\_reweight}, \quad (3)$$

$$L_{self\_adv\_reweight} = \sum_i \sum_{p_j \in S_i} w_{i,j} * BCELoss(\sigma(q_i \cdot p_j), 0), \quad (4)$$

$$w_{i,j} = act(q_i \cdot p_j). \quad (5)$$

We experiment with various activation functions $act()$ including identity, sigmoid and ReLU functions. Identity activation function gives the best performance.[6]

**Self-Adversarial Negative Sampling:** Another way of applying the self-adversarial learning is to take the similarity score as the sampling probability:

$$L_{rel} = L_{pos} + \lambda_{neg} * L_{self\_adv\_sampling}, \quad (6)$$

$$L_{self\_adv\_sampling} = \sum_i \sum_{p_j \in S_i} BCELoss(\sigma(q_i \cdot p_j), 0), \quad (7)$$

$$P(p_j \in S_i) = \begin{cases} softmax(w_{i,j}/T) * (T^2), & i \neq j, \\ 0, & i = j, \end{cases} \quad (8)$$

where $T$ is the self-adversarial sampling temperature and $w_{i,j}$ is the similarity score as in Equation 5.

## 5 SYSTEM DEPLOYMENT

We use the trained embedding model for retrieval and ranking. We utilize an approximate nearest neighbor search (FAISS[7]) for retrieval. We use similarity scores of the query and product embeddings as one of the features for our ranking model. Figure 4 demonstrates the overall system architecture.

[6]Details can be found in Appendix B.

### 5.1 Daily Offline Computation Pipeline

Once we train the embedding model, we pre-compute product embeddings, train the ANN indices offline daily. At Instacart, we have thousands of retailers and they don't share the same product set, and users first choose a retailer they want to shop at. [7] This poses different options for building the ANN indices. A straightforward way is to build a single index containing all the products available. It is simple and requires less memory since there are no duplicated embeddings in the index. This option cannot guarantee the retrieved products belong to the retailer that the users choose. We could retrieve a large number of products and do a post filtering, but this increases search time.

Another option is to build an individual index for each retailer. We can avoid retrieving products from irrelevant retailers, but the overall ANN model size will increase due to repeated products. The total size of all the indices is less than 4x of the combined index enough to be fitted in a single server instance, so we choose to build per-retailer ANN indices. The pre-computed product embeddings are also indexed into the catalog store daily, so they can be used during online serving.

### 5.2 Online Serving

During serving time, we compute the query embedding with the query encoder and use the ANN service to retrieve top $k$ nearest neighbors in the product embeddings set.

Embedding based retrieval (EBR) is efficient especially for non-keyword matching cases which will be discussed in later Section 7.2. However, EBR can include irrelevant products occasionally if we use the top $k$ method. To address this challenge, we apply the following rules.

- First we retrieve top $k$ products from the ANN service.
- We drop products if the similarity scores are below a certain threshold regardless of the ranking.
- We construct a white list categories based on the most top several products, and drop the products falling outside the white list categories.
- We apply availability filtering based on warehouse data.
- We finally merge the retrieved products with the keyword-based retrieval results.

We return the similarity scores for all the products regardless of the retrieval source by using the pre-indexed product embedding in the catalog. We perform final re-ranking for the merged retrieved products with similarity score as a feature.

## 6 EXPERIMENTS

### 6.1 Training Details

We base all our experiments on a pre-trained language model. We choose to use MiniLM-L3-v2[8] since it achieves relatively good performance and also provide the fastest speed, which can be beneficial for online inference.

[7]We also allow users to search first before they choose a retailer, which we call cross retailer search. We provide a list of retailers given a query, once users click one of the retailers from the list, we send a request to retrieve items from ANN service.
[8]https://huggingface.co/sentence-transformers/paraphrase-MiniLM-L3-v2

|  | NDCG@5 | NDCG@10 | NDCG@20 | RECALL@5 | RECALL@10 | RECALL@20 |
|---|---|---|---|---|---|---|
| BM25 [16] | 0.5413 | 0.5785 | 0.6389 | 0.6365 | 0.6372 | 0.6371 |
| Wide & Deep [2] | 0.7609 | 0.7714 | 0.7936 | 0.8326 | 0.8112 | 0.7955 |
| Que2Search [10] | **0.8158** | 0.8242 | 0.8367 | 0.8938 | 0.8772 | 0.8650 |
| Proposed | 0.8150 | **0.8246** | **0.8372** | **0.9013** | **0.8863** | **0.8751** |

Table 1: Offline Evaluation Results. Our proposed method outperforms various baseline models.

We initialize both the query and the product encoders with the MiniLM-L3 model. As described in Section 3, we train the model with two steps. In the best setting, We train the model with shared parameters in both towers in the first warm-up training step. We train the model on the "warm-up dataset" with binary classification loss, using all other in-batch products as negative samples. In the second cascade training step, we continue training the two towers but with un-tied parameters. We train the model on the "cascade training dataset" with self-adversarial negative re-weighting training strategy. More training details can be found in Appendix A

### 6.2 Offline Evaluation Metrics

We create an offline evaluation dataset to measure the model performance so we don't have to launch online testing for every training trial. The user engagement log is not an ideal testing distribution because: first, it comes from the same source as the training data; second, it just reflects the conversion relationships instead of relevance representation ability; and third, when evaluating on the test data collected from the user engagement log, we get 0.999 NDCG@5 for most training trials.

Other than the previous proposed evaluation metric RECALL@K in [5, 10], we propose to also monitor NDCG@K on a human evaluation dataset as offline evaluation. We collect sampled search log to human annotators, query product pairs are labeled into 5 categories:

- **Strongly relevant**: The product is exactly the (type of) product the query is looking for.
- **Relevant**: The product is the type of the product the query is looking for, but there are likely others that fit better.
- **Somewhat relevant**: The product is not exactly what the query is looking for but I understand why it was shown.
- **Not Relevant**: The product is not what the query is looking for, and I can't imagine why it was shown.
- **Offensive**: The product is unacceptable and creates a bad experience.

After filtering out records that raters do not achieve agreement on, meaning 3 out of 5 raters gives the same rate, we got 158k query product pairs. We take "strongly relevant" as a score of 3, "relevant" as 2, "somewhat relevant" as 1 and the other levels as zero. We then compute NDCG@K based on these scaled scores. Also, we take the levels with positive scores as positive labels and compute RECALL@K as other metrics.

### 6.3 Offline Evaluation Results

We show how the proposed method outperforms various baseline models in table 1. We include BM25 [16] algorithm, Wide & Deep [2] model (our previous production model), and Que2Search [10] as baseline models. Wide & Deep uses historical conversion features and text matching scores as features for the wide component, and product raw text data and query text as features for the deep component. The deep component is LSTM-based [4] network followed by 2 stage dense layers. For Que2Search [10], we tried symmetrical scaled cross entropy loss in the 1st stage and applied margin ranking loss in the 2nd stage training. We performed extensive hyper parameter search for fair comparison. We tried *scale* between 15 and 20 for the symmetrical cross entropy loss, and *margin* between 0.1 and 0.2 for the margin ranking loss as in [10]. We didn't observe improvement using margin ranking loss, thus we only report the best results obtained with the symmetrical scaled cross entropy loss.

BM25 performs worst compared to any other models which shows the effectiveness of representation learned from the deep learning models. Compared with the previous production model (Wide & Deep), the proposed method has shown significant improvement. Que2Search shows pretty good results, but our proposed method outperforms Que2Search especially for RECALL@K metrics that are critical when used in retrieval.

### 6.4 Ablation Study

We study the effect of proposed methods using the ablation study. The results are shown in table 2.

**The Self-Adversarial Negative Learning Method Helps:** Comparing row *Vanilla Embedding* and row *+Self-Adv*, we can see the proposed self-adversarial method can boost the model's performance, especially on ranking metrics (NDCG).

Comparing row *Vanilla Embedding* and *+Data* in Table 1, we find vanilla all in-batch negative training can benefit from adding more data, by changing the training set from the the cascade training set (6M) to the warm-up set (14M). However, the model could not gain further benefit while we combine with a self-adversarial training strategy, comparing row *+Self-Adv* and *+Data&+Self-Adv*. This is because although the warm-up dataset is larger, it is also nosier, thus is not an ideal training ground truth.

**Effect of cascade training:** After adding the cascade training, row *+Cascade* outperforms previous rows with a large gap. Another interesting point is that row *+Cascade\** outperforms row *+Cascade*. The second cascade training experiment first makes full use of all the in-batch negative samples at the warm-up stage and then learns more accurate information in the cascade training stage through self-adversarial negative re-weighting.

**Effect of cross architecture cascade training:** We also observe that two tower model architecture in the cascade training stage gives better performance than Siamese network architecture, see row *+Two Tower*. In the cascade training stage, further fine-tuning

| | 2-Tower | Data & Negative | NDCG@5 | NDCG@10 | NDCG@20 | RECALL@5 | RECALL@10 | RECALL@20 |
|---|---|---|---|---|---|---|---|---|
| Vanilla Embedding | N | Cas | 0.8051 | 0.8160 | 0.8301 | 0.8855 | 0.8692 | 0.8574 |
| +Data | N | Warm | 0.8121 | 0.8215 | 0.8347 | 0.8915 | 0.8756 | 0.8638 |
| +Self-Adv | N | Cas(Adv) | 0.8151 | 0.8232 | 0.8362 | 0.8921 | 0.8748 | 0.8623 |
| +Data & +Self-Adv | N | Warm(Adv) | 0.8074 | 0.8181 | 0.8319 | 0.8880 | 0.8724 | 0.8607 |
| +Cascade | N | Warm(Adv)→Cas(Adv) | 0.8102 | 0.8210 | 0.8342 | 0.8989 | 0.8845 | 0.8728 |
| +Cascade* | N | Warm→Cas(Adv) | 0.8139 | 0.8242 | 0.8367 | 0.9006 | 0.8853 | 0.8739 |
| +Two Tower | Y | Warm→Cas(Adv) | 0.8150 | **0.8246** | **0.8372** | **0.9013** | **0.8863** | **0.8751** |

**Table 2: For negative sampling, if not specified, we use all in-batch negative samples. If labeled as *Adv*, we are applying the self-adversarial negative re-weighting strategy. *Warm* and *Cas* in the Data & Negative column means warm-up and cascade training dataset respectively.**

| | MRR | CAPS | GMV |
|---|---|---|---|
| Wide & Deep [2] | - | - | - |
| + Vanilla Embedding | +0.7% | +2.0% | +0.5% |
| + Proposed | +1.2% | +4.1% | +1.5% |

**Table 3: Online A/B testing results. MRR: Mean Reciprocal Rank of the first converted item, CAPS: Cart Adds Per Search, GMV: Gross Merchandise Value**

model parameters individually makes the model to efficiently learn customized information for each encoder.

## 6.5 Online Evaluation Results

We deploy our system to production and conduct A/B testing to prove the effectiveness of the proposed methods. In Table 3 we show the relative improvement of the proposed methods compared with our baseline wide and deep model. When we use the embedding model trained with vanilla in-batch negative samples (Vanilla Embedding in table 3), we observe 0.7 % improvement in MRR, 2.0% gain in CAPS, and 0.5% of GMV increase. With all the proposed methods applied, we achieve 1.5 % improvement in MRR, 4.1% gain in CAPS, which turns into 1.2% of GMV increase.

## 7 ANALYSIS

### 7.1 Semantic Relevance

To analyze how the embedding model performs, we visualize the product embedding vectors. We randomly sampled product embeddings and colored them according to their pre-labeled categories. We apply T-SNE [19] to project the embedding vectors into a 2D space. As shown in Figure 5, products with the same category labels are well clustered, and different clusters are well separated with a large margin.

We also plot the model predicted similarity score distribution on our human evaluation dataset in Figure 6. We color samples based on the human rated relevance levels. As can be seen, higher relevance groups are distributed to the right side, which means higher relevance score predictions. The non-relevant group is located at the most left area, with a much lower relevance score, separated from the relevant groups. This distribution shows the proposed method can help predict the semantic relevance very well.

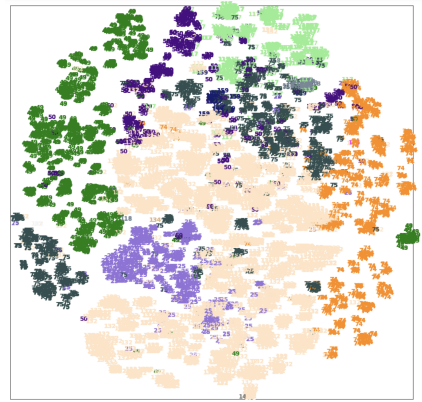

**Figure 5: Embedding Visualization. Colored by different product categories. We sampled products from *food* and the light orange color represents the sub-category *pantry*.**

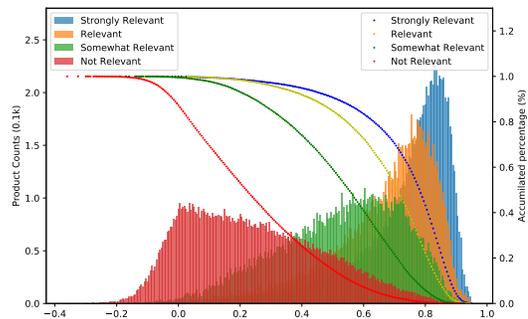

**Figure 6: Score distribution over the human evaluation dataset.**

### 7.2 Learned Semantic Signals outperforms Historical Engagement Signals

We first show the learned representation performs well with an example query: "Milk". Users usually send this query for products under our "Plain Milk" category. We compare the scores under this category with another category "Milk Chocolate", which also contains the word "Milk" and might cause failure in keyword based systems. We collect the click through rates (CTRs) and the embedding scores for randomly selected products under both categories given "milk" as a query. Figure 7(a) shows CTRs and embedding scores in descending order. From the figure, we observe that almost all the products in the "Plain Milk" category have higher embedding scores than "Milk Chocolate" products. That means embedding score can be used as a good indicator for differentiating between "Plain Milk" products and "Milk Chocolate" products. Half of the "Plain Milk" products have zero CTRs and some "Milk Chocolate" products have non-zero CTRs. If we rank items with CTRs, half of the relevant milk products get ranked lower than several milk chocolate products. That shows CTR is not a good search relevance indicator.

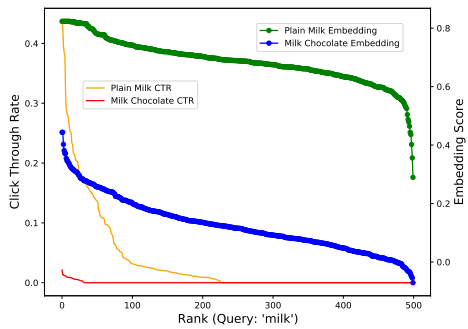

(a) Score Distribution.

**Figure 7: Comparing Embedding Scores with Click Through Rate for query "milk".**

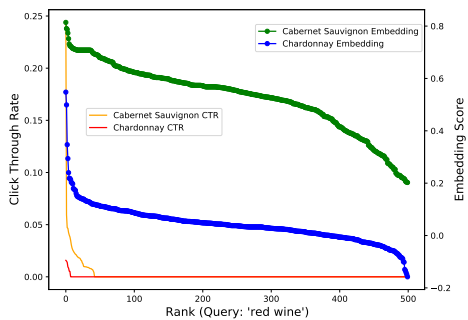

(a) Score Distribution.

**Figure 8: Comparing Embedding Scores with Click Through Rate for query "red wine".**

We then show how the proposed methods perform on the cold-start products with the example query "red wine". There are hundreds of different red wines in our product set, and very few of them are purchased through search. We compare similarity scores for products under "Cabernet Sauvignon" (one of red wine categories) and "Chardonnay" (one of white wine categories). We collect the CTRs and the embedding scores for randomly selected products under the two categories given "red wine" as a query. There are only about 10 % products under the "Cabernet Sauvignon" with positive CTRs, meaning all the other products are cold but relevant products. Figure 8(a) shows that embedding scores can be used for separating products in "Cabernet Sauvignon" from the products in "Chardonnay" while CTRs suffer from the cold start problem.

## 8 APPLICATIONS

We envision that the search embedding model can be used as a component for other downstream tasks.

**Ads relevance model:** The search embedding model is used as a checkpoint for fine-tuning ads relevance models at Instacart. The Ads relevance has different application scenarios but can be easily transferred from the search relevance model.

**Semantic deduplication for auto suggestion:** To increase the diversity of auto suggestion in the search bar, we perform deduplication on the candidate query set using the embedding model: If the similarity score of two candidate queries is very high, we remove one of them in auto suggestion.

**Semantic query clustering:** Historical features like click through rate (CTR) are important features in our search ranking. We observe that engagement data used for CTR calculation are highly imbalanced across queries, while more user engagement log usually means more accurate information. We apply embedding based similarity scores to merge log data for queries that are semantically similar. By sharing the user engagement data, less popular queries can perform equally well as their more popular neighboring queries.

## 9 CONCLUSION

In this paper, we propose to use only content-based features, and add catalog synthesis data to help with cold starts. We propose cross architecture cascade training, as well as self-adversarial negative sampling/re-weighting, to efficiently train the model on the noisy data. The proposed model outperforms the previous baseline by 10% in RECALL@20 on an offline human evaluation dataset. In online A/B testing, the new system improves CAPS and GMV by 4.1% and 1.5% respectively. We share our knowledge in model architecture, training details and system deployment. We also share our insights on how to further utilize the embedding model in various applications including ads relevance model, auto suggestion, and click through rate (CTR) computation.

## REFERENCES

[1] Tom B. Brown, Benjamin Mann, Nick Ryder, Melanie Subbiah, Jared Kaplan, Prafulla Dhariwal, Arvind Neelakantan, Pranav Shyam, Girish Sastry, Amanda Askell, Sandhini Agarwal, Ariel Herbert-Voss, Gretchen Krueger, Tom Henighan, Rewon Child, Aditya Ramesh, Daniel M. Ziegler, Jeffrey Wu, Clemens Winter, Christopher Hesse, Mark Chen, Eric Sigler, Mateusz Litwin, Scott Gray, Benjamin Chess, Jack Clark, Christopher Berner, Sam McCandlish, Alec Radford, Ilya Sutskever, and Dario Amodei. 2020. Language Models are Few-Shot Learners. In *Advances in Neural Information Processing Systems 33: Annual Conference on Neural Information Processing Systems 2020, NeurIPS 2020, December 6-12, 2020, virtual*, Hugo Larochelle, Marc'Aurelio Ranzato, Raia Hadsell, Maria-Florina Balcan, and Hsuan-Tien Lin (Eds.).
[2] Heng-Tze Cheng, Levent Koc, Jeremiah Harmsen, Tal Shaked, Tushar Chandra, Hrishi Aradhye, Glen Anderson, Greg Corrado, Wei Chai, Mustafa Ispir, Rohan Anil, Zakaria Haque, Lichan Hong, Vihan Jain, Xiaobing Liu, and Hemal Shah. 2016. Wide & Deep Learning for Recommender Systems. https://doi.org/10.48550/ARXIV.1606.07792

[3] Jacob Devlin, Ming-Wei Chang, Kenton Lee, and Kristina Toutanova. 2019. BERT: Pre-training of Deep Bidirectional Transformers for Language Understanding. In *Proceedings of the 2019 NAACL*. ACL, Minneapolis, Minnesota, 4171–4186.

[4] Sepp Hochreiter and Jürgen Schmidhuber. 1997. Long short-term memory. *Neural computation* 9, 8 (1997), 1735–1780.

[5] Jui-Ting Huang, Ashish Sharma, Shuying Sun, Li Xia, David Zhang, Philip Pronin, Janani Padmanabhan, Giuseppe Ottaviano, and Linjun Yang. 2020. Embedding-based Retrieval in Facebook Search. *Proceedings of the 26th ACM SIGKDD*.

[6] Xiaoqi Jiao, Yichun Yin, Lifeng Shang, Xin Jiang, Xiao Chen, Linlin Li, Fang Wang, and Qun Liu. 2020. TinyBERT: Distilling BERT for Natural Language Understanding. In *Findings of the Association for Computational Linguistics: EMNLP 2020, Online Event, 16-20 November 2020 (Findings of ACL, Vol. EMNLP 2020)*, Trevor Cohn, Yulan He, and Yang Liu (Eds.). ACL, 4163–4174.

[7] Jeff Johnson, Matthijs Douze, and Hervé Jégou. 2017. Billion-scale similarity search with GPUs. *CoRR* abs/1702.08734 (2017). arXiv:1702.08734

[8] Mike Lewis, Yinhan Liu, Naman Goyal, Marjan Ghazvininejad, Abdelrahman Mohamed, Omer Levy, Veselin Stoyanov, and Luke Zettlemoyer. 2020. BART: Denoising Sequence-to-Sequence Pre-training for Natural Language Generation, Translation, and Comprehension. In *Proceedings of the 58th ACL, Online, July 5-10, 2020*, Dan Jurafsky, Joyce Chai, Natalie Schluter, and Joel R. Tetreault (Eds.). ACL, 7871–7880.

[9] Sen Li, Fuyu Lv, Taiwei Jin, Guli Lin, Keping Yang, Xiaoyi Zeng, Xiao-Ming Wu, and Qianli Ma. 2021. Embedding-Based Product Retrieval in Taobao Search. In *Proceedings of the 27th ACM SIGKDD* (Singapore) *(KDD '21)*. ACM, New York, NY, USA, 3181–3189.

[10] Yiqun Liu, Kaushik Rangadurai, Yunzhong He, Siddarth Malreddy, Xunlong Gui, Xiaoyi Liu, and Fedor Borisyuk. 2021. Que2Search: Fast and Accurate Query and Document Understanding for Search at Facebook. In *Proceedings of the 27th ACM SIGKDD* (Singapore) *(KDD '21)*. ACM, New York, NY, USA, 3376–3384.

[11] Priyanka Nigam, Yiwei Song, Vijai Mohan, Vihan Lakshman, Weitian Ding, Ankit Shingavi, Choon Hui Teo, Hao Gu, and Bing Yin. 2019. Semantic Product Search. *CoRR* abs/1907.00937 (2019). arXiv:1907.00937

[12] Priyanka Nigam, Yiwei Song, Vijai Mohan, Vihan Lakshman, Weitian Allen Ding, Ankit Shingavi, Choon Hui Teo, Hao Gu, and Bing Yin. 2019. Semantic Product Search. In *Proceedings of the 25th ACM SIGKDD, Anchorage, AK, USA, August 4-8, 2019*, Ankur Teredesai, Vipin Kumar, Ying Li, Rómer Rosales, Evimaria Terzi, and George Karypis (Eds.). ACM, 2876–2885.

[13] Xiao Qin, Nasrullah Sheikh, Berthold Reinwald, and Lingfei Wu. 2021. Relation-aware Graph Attention Model with Adaptive Self-adversarial Training. In *Thirty-Fifth AAAI Conference on Artificial Intelligence, AAAI 2021, Thirty-Third Conference on Innovative Applications of Artificial Intelligence, IAAI 2021, The Eleventh Symposium on Educational Advances in Artificial Intelligence, EAAI 2021, Virtual Event, February 2-9, 2021*. AAAI Press, 9368–9376.

[14] Colin Raffel, Noam Shazeer, Adam Roberts, Katherine Lee, Sharan Narang, Michael Matena, Yanqi Zhou, Wei Li, and Peter J. Liu. 2020. Exploring the Limits of Transfer Learning with a Unified Text-to-Text Transformer. *J. Mach. Learn. Res.* 21 (2020), 140:1–140:67.

[15] Nils Reimers and Iryna Gurevych. 2019. Sentence-BERT: Sentence Embeddings using Siamese BERT-Networks. In *Proceedings of the 2019 EMNLP-IJCNLP*. ACL, Hong Kong, China, 3982–3992.

[16] Stephen Robertson and Hugo Zaragoza. 2009. The probabilistic relevance framework: BM25 and beyond. *Foundations and Trends in Information Retrieval* 3, 4 (2009), 333–389.

[17] Victor Sanh, Lysandre Debut, Julien Chaumond, and Thomas Wolf. 2020. DistilBERT, a distilled version of BERT: smaller, faster, cheaper and lighter. arXiv:1910.01108 [cs.CL]

[18] Zhiqing Sun, Zhi-Hong Deng, Jian-Yun Nie, and Jian Tang. 2019. Rotate: Knowledge graph embedding by relational rotation in complex space. *arXiv preprint arXiv:1902.10197* (2019).

[19] Laurens van der Maaten and Geoffrey Hinton. 2008. Visualizing Data using t-SNE. *Journal of Machine Learning Research* 9 (2008), 2579–2605.

[20] Alex Wang, Amanpreet Singh, Julian Michael, Felix Hill, Omer Levy, and Samuel R. Bowman. 2019. GLUE: A Multi-Task Benchmark and Analysis Platform for Natural Language Understanding. arXiv:1804.07461 [cs.CL]

[21] Wenhui Wang, Furu Wei, Li Dong, Hangbo Bao, Nan Yang, and Ming Zhou. 2020. MiniLM: Deep Self-Attention Distillation for Task-Agnostic Compression of Pre-Trained Transformers. In *Advances in Neural Information Processing Systems 33: Annual Conference on Neural Information Processing Systems 2020, NeurIPS 2020, December 6-12, 2020, virtual*, Hugo Larochelle, Marc'Aurelio Ranzato, Raia Hadsell, Maria-Florina Balcan, and Hsuan-Tien Lin (Eds.).

[22] Xin Wang, Yudong Chen, and Wenwu Zhu. 2021. A Survey on Curriculum Learning. arXiv:2010.13166 [cs.LG]

[23] John Wieting, Mohit Bansal, Kevin Gimpel, and Karen Livescu. 2016. Towards Universal Paraphrastic Sentence Embeddings. In *4th International Conference on Learning Representations, ICLR 2016, San Juan, Puerto Rico, May 2-4, 2016, Conference Track Proceedings*, Yoshua Bengio and Yann LeCun (Eds.).

[24] Tao Wu, Ellie Ka In Chio, Heng-Tze Cheng, Yu Du, Steffen Rendle, Dima Kuzmin, Ritesh Agarwal, Li Zhang, John R. Anderson, Sarvjeet Singh, Tushar Chandra, Ed H. Chi, Wen Li, Ankit Kumar, Xiang Ma, Alex Soares, Nitin Jindal, and Pei Cao. 2020. Zero-Shot Heterogeneous Transfer Learning from Recommender Systems to Cold-Start Search Retrieval. *CoRR* abs/2008.02930 (2020). arXiv:2008.02930

[25] Zhilin Yang, Zihang Dai, Yiming Yang, Jaime Carbonell, Ruslan Salakhutdinov, and Quoc V. Le. 2019. XLNet: Generalized Autoregressive Pretraining for Language Understanding. In *Proceedings of the 33rd International Conference on Neural Information Processing Systems*. Curran Associates Inc., Red Hook, NY, USA.

[26] Wen-tau Yih, Kristina Toutanova, John C. Platt, and Christopher Meek. 2011. Learning Discriminative Projections for Text Similarity Measures. In *Proceedings of the 15th Conference on Computational Natural Language Learning*. ACL, Portland, Oregon, USA, 247–256.

[27] Han Zhang, Songlin Wang, Kang Zhang, Zhiling Tang, Yunjiang Jiang, Yun Xiao, Weipeng Yan, and Wenyun Yang. 2020. Towards Personalized and Semantic Retrieval: An End-to-End Solution for E-commerce Search via Embedding Learning. In *Proceedings of the 43rd International ACM SIGIR, China, July 25-30, 2020*, Jimmy Huang, Yi Chang, Xueqi Cheng, Jaap Kamps, Vanessa Murdock, Ji-Rong Wen, and Yiqun Liu (Eds.). ACM, 2407–2416.

## A TRAINING DETAILS

We experiment with different pre-trained models including distilbert [9], MiniLM [10], TinyBERT [11]. We also experiment with their semantically fine-tuned checkpoints, for example the ones fine-tuned on sts-b or QQP in GLUE [20], or other NLI and paraphrase datasets.

Our implementation is based on the sentence transformer [15][12]. The base model outputs a hidden vector for each input sentence. We project the hidden vector into a lower (100) dimension with dense layers to meet the memory requirements. We randomly initialize all other model parameters.

We use grid search to find the best hyper parameters:

- Training Epochs: 1,2,3,**4**,5,6;
- Warmup Percentage: 0%, 5% , **8%**, 10% 20%, 50%;
- Learning Rate: 1e-5,4e-5, 8e-5, **1.6e-4**, 3.2e-4, 1.2e-3;
- Batch Size: we fix it to **512**, the largest allowed size;

We train the model with Adam optimizer, and amp (automatic mixed precision) enabled. We conduct our experiments on Amazon ml.p3.8xlarge EC2 instances, which contains 64G V100 GPUs .

## B DIFFERENT ACTIVATION FOR SELF-ADVERSARIAL NEGATIVE SAMPLING.

Similar to Section 7.1, we plot the score distribution on our human evaluation dataset in Figure 9. We group different query and product pairs based on the human labels, then we plot their score distributions. We experiment with different activation functions for self-adversarial negative re-weighting as explained in Section 4.3.

- **Identity activation:** $w_{i,j} = q_i \cdot p_j$;
- **Sigmoid acitvation:** $w_{i,j} = \frac{1}{1+e^{-q_i \cdot p_j}}$;
- **ReLU acitvation:** $w_{i,j} = \max(q_i \cdot p_j, 0)$.

As can be seen from the Figure 9, using all proposed activation functions can somehow separate the four categories of relevance items in distribution. However, we can find the identity activation function is giving the best separation effect: the not relevant group is clustered more to the low-relevance score end, while the other three categories are distributed to the right side, which means higher relevance scores.

## C EFFECT OF THE TWO-TOWER MODEL

We compare different results when we turn the model in to a two-tower fashion. We can see, for each group in Table 4, the model all shows a better performance when we separate the encoder parameters in query and product towers.

## D ABLATION STUDY ON SELF-ADVERSARIAL NEGATIVE STRATEGIES.

In Section 4.2 and 4.3 we introduced different methods of negative sampling. Table 5 shows the detailed experiment result. We note different experiments according to their different negative sample strategies:

---

[9]https://huggingface.co/distilbert-base-uncased
[10]https://huggingface.co/sentence-transformers/paraphrase-MiniLM-L3-v2
[11]https://huggingface.co/cross-encoder/stsb-TinyBERT-L-4
[12]https://www.sbert.net/

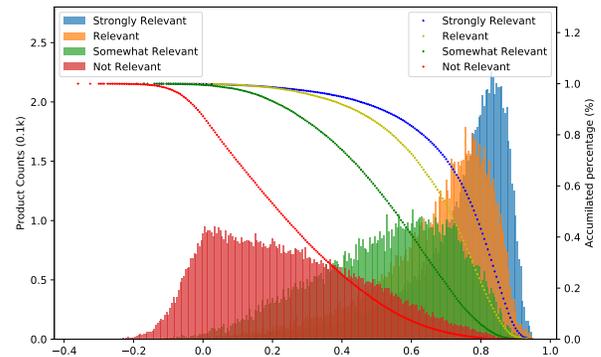

(a) Identity Adversarial.

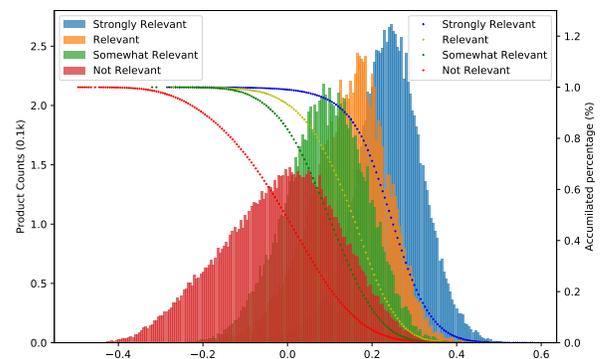

(b) Sigmoid Adversarial.

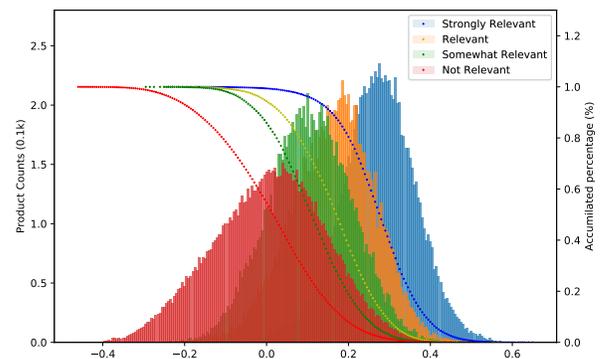

(c) ReLU Adversarial

**Figure 9: Score distribution over different categories in the human evaluation dataset. We compare three self-adversarial re-weighting activation functions: identity function and sigmoid function.**

- **All in-batch negative**: we include all the in-batch negative samples in the loss.
- **Random - $k$**: we randomly sampled $k$ in-batch negative samples from a batch size of 512.

| 2-Tower | Data | RECALL@5 | RECALL@10 | RECALL@20 |
|---|---|---|---|---|
| N | Cas | 0.8855 | 0.8692 | 0.8574 |
| N | Warm(Adv)→Cas(Adv) | 0.8989 | 0.8845 | 0.8728 |
| Y | Warm(Adv)→Cas(Adv) | 0.8996 | 0.8850 | 0.8739 |
| N | Warm→Cas(Adv) | 0.9006 | 0.8853 | 0.8739 |
| Y | Warm→Cas(Adv) | 0.9013 | 0.8863 | 0.8751 |

Table 4: Ablation study for two-tower architecture.

| Negative Sampling | NDCG@5 | NDCG@10 | NDCG@20 | RECALL@5 | RECALL@10 | RECALL@20 |
|---|---|---|---|---|---|---|
| Baseline (All In-batch Negative) | 0.8051 | 0.8160 | 0.8301 | 0.8855 | 0.8692 | 0.8574 |
| Random - 1 | 0.8086 | 0.8189 | 0.8322 | 0.8871 | 0.8713 | 0.8595 |
| Random - 16 | 0.8113 | 0.8210 | 0.8343 | 0.8907 | 0.8746 | 0.8627 |
| Self-adv Re-weight | 0.8083 | 0.8198 | 0.8332 | **0.9004** | **0.8862** | **0.8746** |
| Self-adv Sampling - all | 0.8051 | 0.8160 | 0.8301 | 0.8855 | 0.8692 | 0.8574 |
| Self-adv Sampling - 16 | **0.8151** | **0.8232** | **0.8362** | 0.8921 | 0.8748 | 0.8623 |

Table 5: For Negative Sampling, trained with shared tower parameters and cascade training dataset.

- **Self-adv Re-weight**: we apply the self-adversarial negative re-weighting method, that multiply all the samples in the batch with their prediction scores as a re-weighting factor.
- **Self-adv Sampling** - $k$: we apply the self-adversarial negative sampling method and sampled $k$ in-batch negative samples from a batch of 512.

From the results in Table 5, we can see that reducing the negative samples included in the loss function to an appropriate size can already help improve the model's performance. Further applying the self-adversarial negative sampling can help boost the training the most. With this mechanism, the model is able to adjust each individual sample's contribution as the training goes. One interesting observation is that, the self-adversarial re-weighting strategy provides more gain to recall while the self-adversarial sampling strategy provides more gain to NDCG. We hypothesize that this is because re-weighting would ignore the degree of hardness between samples by punishing more on the hard negatives so that they are classified correctly (positive or negative). However, the re-sampling strategy doesn't punish more on the hard negatives, thus the model would learn the different level of relevance, which results in better performance in ranking metrics.